# A Single Channel-Based Neonatal Sleep-Wake Classification using Hjorth Parameters and Improved Gradient Boosting

Muhammad Arslan[1], Muhammad Mubeen[2], Saadullah Farooq Abbasi[3], Muhammad Shahbaz Khan[4], Wadii Boulila[5], Jawad Ahmad[4]

[1]Department of Computer Science, Loyola University, Chicago Illinois 60660, United States
[2] Department of Computer Science, University of People, Pasadena, CA 91101, United States
[3]Department of Electronic, Electrical and Systems Engineering, University of Birmingham, Birmingham B15 2TT, United Kingdom
[4]School of Computing, Engineering and the Built Environment, Edinburgh Napier University, Edinburgh EH105DT, United Kingdom
[5]RIOTU Lab, Prince Sultan University, Riyadh 12435, Saudi Arabia

**Abstract**

Sleep plays a crucial role in neonatal development. Monitoring the sleep patterns in neonates in a Neonatal Intensive Care Unit (NICU) is imperative for understanding the maturation process. While polysomnography (PSG) is considered the best practice for sleep classification, its expense and reliance on human annotation pose challenges. Existing research often relies on multichannel EEG signals; however, concerns arise regarding the vulnerability of neonates and the potential impact on their sleep quality. This paper introduces a novel approach to neonatal sleep stage classification using a single-channel gradient boosting algorithm with Hjorth features. The gradient boosting parameters are fine-tuned using random search cross-validation (randomsearchCV), achieving an accuracy of 82.35% for neonatal sleep-wake classification. Validation is conducted through 5-fold cross-validation. The proposed algorithm not only enhances existing neonatal sleep algorithms but also opens avenues for broader applications.

**Introduction**

Sleep plays an important role in neonatal life, serving as a fundamental requirement for both brain and physical development [1]. Inadequate sleep has been associated to brain immaturities, poor emotional regulation, compromised overall wellbeing, and other associated conditions [2]. The initial phase in understanding brain development involves the classification of sleep-wake cycling (CWC) [3]. Fetal sleep is considered a primary driver of neural activity, essential for critical processes such as neuronal survival, and synapse maturation [4]. Nevertheless, within the Neonatal Intensive Care Unit (NICU), newborns encounter various external stimuli that can markedly influence their patterns of sleep and wakefulness [5]. There are two important ways in which continuous monitoring of newborn sleep in the NICU may enhance care. Firstly, since newborn sleep is important for development, monitoring it can serve as a biomarker to predict outcomes by indicating an infant's maturity level and the severity of illness [6]. Secondly, real-time sleep monitoring enables adaptive care tailored to sleep stages, promoting optimal brain development in neonates [7].

The current best practice for classifying neonatal sleep is polysomnography (PSG), which extracts multi-biophysiological signals, including electroencephalography (EEG), electrocardiography (ECG), electrooculography (EOG), and respiration [8]. Despite being the standard practice in contemporary sleep centres, where sleep technicians manually perform it for identifying sleep stages, PSG's manual approach is both time-consuming and expensive. To alleviate labour costs, extensive research has been conducted to identify key features that can automate the classification of sleep stages. In addition to PSG, various techniques have been employed using multi-channel electroencephalogram (EEG) signals [9] as feature sets for sleep classification algorithms. Machine learning (ML) techniques, including ensemble learning and support vector machines (SVMs), have been explored for this purpose [10, 11]. For instance, Fraiwan et al. [12] present an algorithm using deep autoencoders, achieving an accuracy of 80.4%; however, only 17% of the wake samples were correctly classified. Addressing this, Abbasi et al. [13] proposed a multilayer perceptron (MLP) neural network in 2020, achieving an accuracy of 82.53% for awake classification. To further enhance the accuracy, Abbasi et al. [10] proposed an ensemble learning algorithm combining MLP, Convolutional neural network (CNN), and SVM, using SVM for ensembling, and achieving notable performance. Despite the success of ML and deep learning (DL)-based algorithms, the fragility of neonates and the potential impact of multichannel EEG extraction on their sleep quality prompted Awais et al. to propose an algorithm using single-channel EEG [14]. This achieved an accuracy of 77.5% with a mean kappa of 0.55. To further improve, Zhu et al. [15] proposed an algorithm with a squeeze-and-excitation block, achieving an accuracy of 75.4%, though awake classification remained a challenge.

This paper, in this regard, proposes a novel and improved sleep-wake classification method using gradient boosting. Thirteen prominent features have been extracted from single-channel neonatal EEG, and the gradient boosting algorithm has been employed for classification. Random search cross-validation was deployed to select the best parameters of gradient boosting. Finally, to validate the overall dataset, 5-fold cross validation has been used.

The rest of the paper is organized in the following manner: Section II presents the materials and methods, Section III presents the results of the gradient boost algorithm, Section IV discusses the overall paper, and finally Section V concludes the paper.

**Materials and Methods:**

This section provides a concise overview of the dataset, preprocessing steps, feature extraction, and the methodology employed in the study. Figure 1 illustrates the experimental setup of the proposed study.

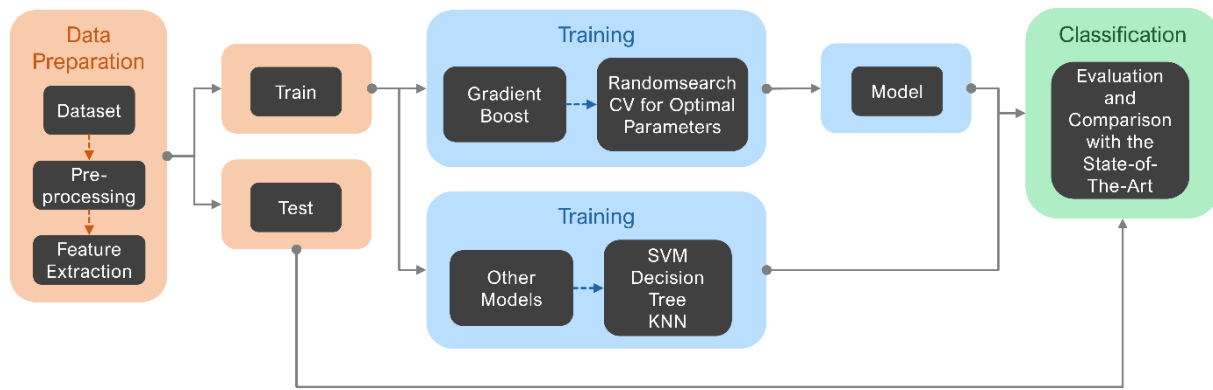

*Figure 1: Flow chart of the proposed algorithm.*

**Dataset:**

Standard Polysomnography (PSG) recordings were obtained from 19 healthy neonates with mild illnesses. Electrodes were positioned following the 10-20 standard electrode placement system [16]. The proposed study recorded nine bipolar EEG channels, one ECG channel, one respiration channel, and one electromyography (EMG) channel. PSG recordings were acquired for a minimum of 2 hours at a sampling rate of 500 Hz, utilizing the NicoletOne EEG system for data extraction and annotation. The flow chart of the proposed study is depicted in the accompanying figure.

**Preprocessing:**

During the recording, EEG data were susceptible to noise and artifacts, including powerline noise (at 50 Hz), baseline noise, and various artifacts such as movement artifact, ECG artifact, and EOG artifact. To mitigate these, a finite impulse response (FIR) filter was designed with a bandpass frequency range of 0.3 Hz to 35 Hz [17]. FIR, known for its linearity, was instrumental in artifact removal. Following noise and artifact removal, the dataset was segmented into 30-second intervals. A total of 4560 segments were extracted from 19 neonates. After segmentation, 1160 segments were excluded from 4560 segments using annotation provided by neurologists (Artifacts). For this study, even if 5% of the segment is annotated as artifact, we excluded that segment. After removing these segments, the study used 3400 segments for training and testing the network.

**Features:**

Given that sleep-wake stages exhibit distinctive patterns in both time and frequency content, EEG analysis encompassed both domains. Time domain features included minima, maxima, mean amplitude, standard deviation (SD), skewness, kurtosis, root mean square of EEG, energy, Hjorth parameters. Frequency domain features comprised spectral centroid, spectral spread, and spectral flatness.

**Hjorth Parameters:**

Hjorth Parameters, widely used statistical indicators in signal processing, were applied to analyze EEG signals within the time realm [18].

**Activity.** The parameter that shows a temporal function's variation and signal strength is called the activity parameter. It can depict the surface of the power spectrum in the frequency domain. The action is shown by equation (1) below:

$$Activity = V(x(t)) \quad (1)$$

Where x(t) reflects the signal and V represents variance.

**Mobility.** The mobility parameter is a representation of the spectrum power's frequency or SD proportion [15]. Mathematically,

$$Mobility = \sqrt{V(dx(t)/d(t))V(x(t))^{-1}} \quad (2)$$

**Complexity.** The variation in frequency is represented by the complexity parameter. Complexity is defined as the ratio of the first derivative of the signal's mobility to the signal's mobility. Equation (3) has the following:

$$Complexity = (\sigma''(\sigma'_s)^{-1})(\sigma'_s(\sigma_s)^{-1})^{-1} \quad (3)$$

**Methodology:**

In the sleep-wake classification scenario, we encounter a system comprising a stochastic "sleep-wake" output variable denoted as $y$ and a set of random input or explanatory variables. The objective, given a training dataset with known sleep-wake values, is to identify a function $f$ that effectively maps the input variables to $y$. The aim is to minimize the expected value of a specified loss function across the joint distribution of all input values.

The boosting technique in this context seeks to approximate the target function $F$ through an "additive" expansion expressed as:

$$F(x) = F_{prev}(x) + \sum_{m=1}^{M} B_m h_m(x; \theta_m) \qquad (4)$$

Here, $h_m$ represents the base learner, chosen as simple functions of the input variables. The expansion coefficients $B_m$ and the parameters $\theta_m$ undergo iterative optimization in a forward "stage-wise" manner. Commencing with an initial guess $F_{prev}$, the model updates at each iteration $m = 1, 2, \ldots, M$ according to:

$$F_m(x) = F_{m-1}(x) + B_m h_m(x; \theta_m) \qquad (5)$$

Gradient boosting approximately resolves the optimization problem through a two-step procedure. Initially, $F_m$ undergoes least squares fit:

$$min_{\beta m, \theta m} \sum_{i=1}^{N} \Psi(y_i, F_{m-1}(x_i) + B_m h_m(x; \theta_m)) \qquad (6)$$

Here, $\Psi$ represents a differentiable loss function. Subsequently, given the current approximation $F_m$, optimal values for $B_m$ and $\theta_m$ are determined. Figure 2 shows the flow of the proposed gradient boosting.

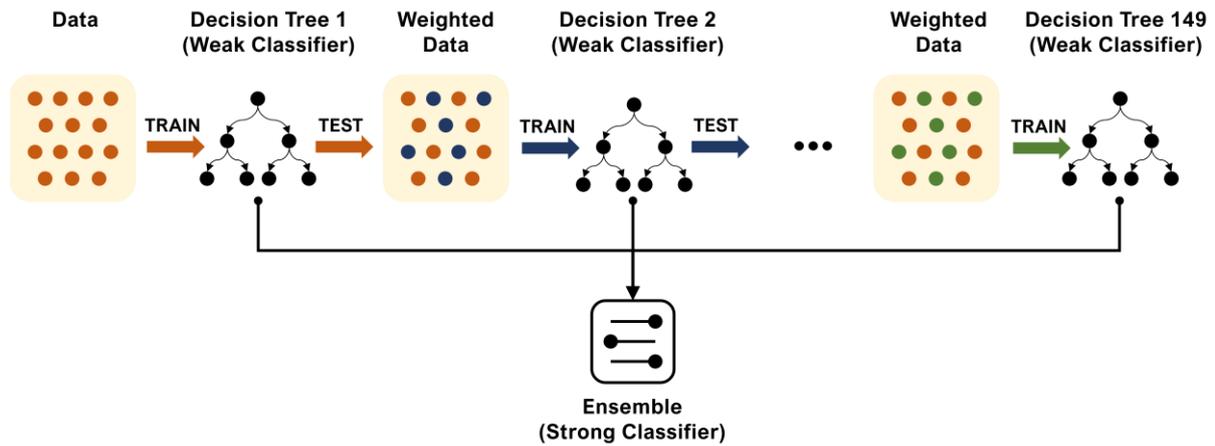

Figure 2: Working flow of gradient boosting tree.

The shrinkage parameter $0 < v \leq 1$ governs the learning rate of the procedure. Empirically, smaller values have proven to yield superior generalization performance in sleep-wake classification. For the proposed study, random search cross validation has been used to select the best parameters for the gradient boosting algorithm i.e., number of estimators, maximum depth, learning rate. The results of the random search cross validation are given in the table 1:

Table 1. Selected parameters after Randomsearch cross validation.

|  | No. of estimators | Max depth | Learning rate |
|---|---|---|---|
| **Best Results** | 149 | 10 | 0.104 |

**Performance evaluation:**

The proposed study was evaluated using multiple performance measures, including Kappa, Sensitivity, Specificity, and Accuracy. The classification's accuracy was mostly measured using the values of True Positive (TP), True Negative (TN), False Positive (FP), and False Negative (FN). The following equations, which represent Sensitivity (Se), Specificity (Sp), Accuracy (Acc), and Kappa, were computed using these metrics as the foundation:

$$Se = \frac{TP}{TP+FN} \times 100\% \qquad (7)$$

$$Sp = \frac{TN}{TN+FP} \times 100\% \qquad (8)$$

$$Acc = \frac{TN+TP}{TN+FN+TP+FP} \times 100\% \qquad (9)$$

$$Kappa = \frac{P_{Agree}-P_{Chance}}{1-P_{Chance}} \qquad (10)$$

Here, TP represents the number of correctly classified awake samples, TN denotes the number of correctly classified sleep samples, and FP and FN represent the number of incorrectly classified awake and sleep samples, respectively. $P_{Agree}$ signifies the percentage of outcomes in agreement between the network and the annotation, while $P_{Chance}$ represents the percentage predicted by chance. The proposed system underwent validation using 5-fold cross-validation, where the neonatal dataset was divided into 5 folds, with one-fold used for testing while the other four were utilized for training.

**Results:**

The proposed GBT architecture was implemented and evaluated using Keras libraries on the Google Colab Pro platform. Figure 3 displays the confusion matrix, while Figure 4 showcases the performance metrics calculated using the corresponding confusion matrix.

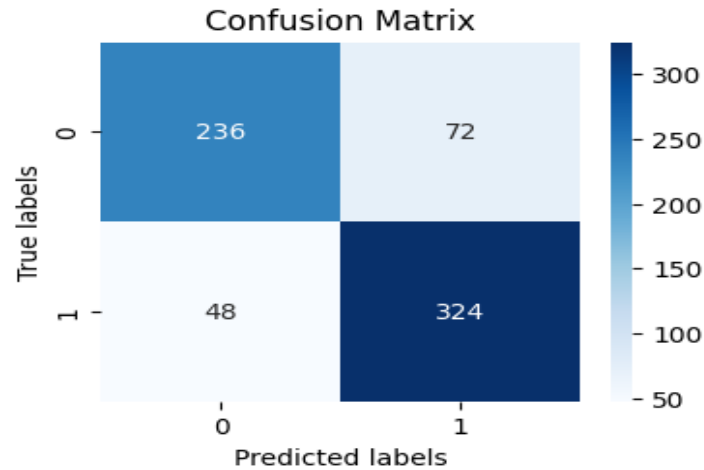

*Figure 3. Confusion matrix of the proposed gradient boosting algorithm.*

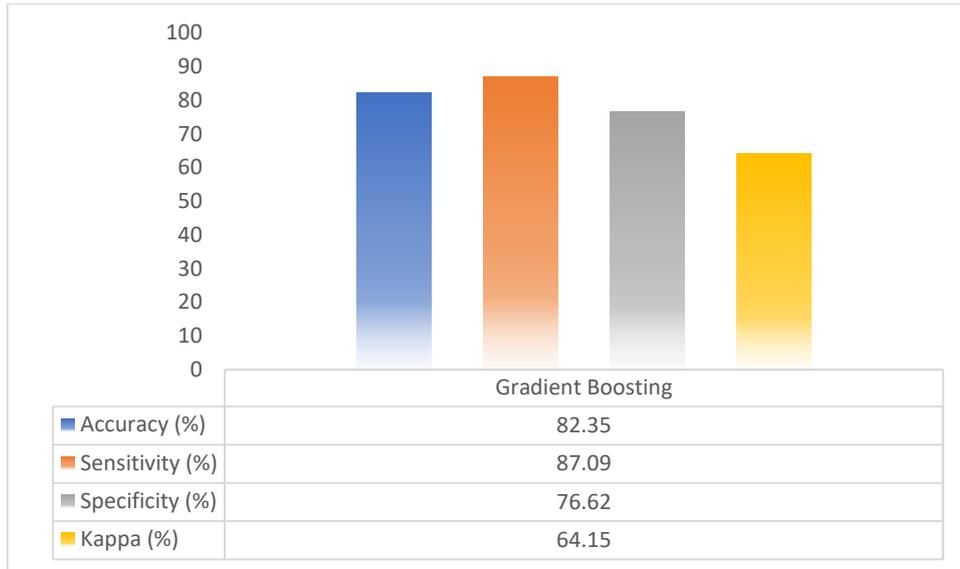

*Figure 4. Performance metrics of the proposed study.*

In the second analysis, when the number of parameters is changed, the accuracy varies based on change in the number of parameters. Similarly, by changes the learning rate of the gradient boosting algorithm, the accuracy of the proposed study decreases significantly.

The classification performance of the proposed Gradient Boosting (GB) is depicted in the ROC curve (Figure 5), illustrating the trade-off between TP rate and FP rate. The area under the mean ROC curve is 91%.

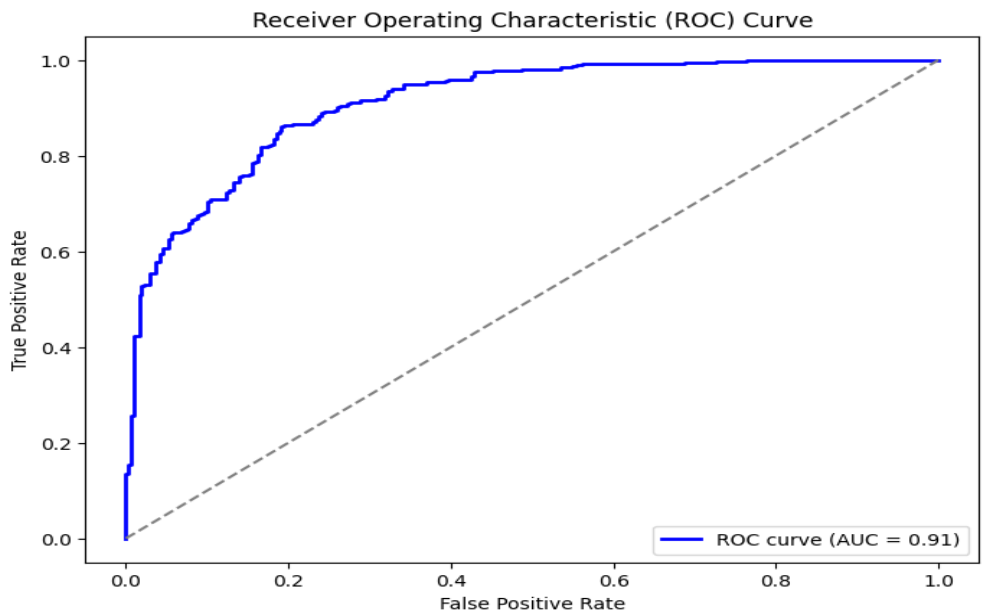

*Figure 5. ROC curve of the proposed gradient boosting algorithm.*

To validate the proposed GB, Table 2 presents a comparative analysis with existing algorithms. The proposed algorithm outperformed all existing algorithms for single-channel classification.

*Table 2. Performance comparison of the proposed study with existing literature.*

|  | Accuracy (%) | Sensitivity (%) | Specificity (%) |
| --- | --- | --- | --- |
| **Proposed study** | **82.35** | **87.09** | **76.62** |
| Gradient Boosting without randomsearchCV | 75.14 | 80.64 | 68.32 |
| SVM | 76.5 | 73 | 78 |
| KNN | 72.5 | - | - |
| Logistic regression | 69.5 | - | - |

Similarly, this study has been juxtaposed with DL algorithms, and the comparative analysis is detailed in Table 3. Various algorithms have previously reported outcomes for the classification of sleep-wake states based on single-channel EEG data. This study conducts a thorough comparison between the results achieved by the proposed Gradient Boosting and those reported by other algorithms, as presented in Table 3.

*Table 3. Performance comparison of the proposed study with deep learning algorithms.*

| | Accuracy (%) | Kappa |
|---|---|---|
| **Proposed Study** | **82.35** | **0.64** |
| MLP | 71 | 0.52 |
| CNN | 69 | - |
| Recurrent Neural network (RNN) | 65 | - |
| Auto Encoders | 17 | - |

**Conclusion**

This paper presented a random search cross validation based gradient boosting algorithm for neonatal sleep-wake classification. The proposed algorithm used 13 time and frequency-domain features, from 3400 30-second EEG samples, for training and testing the algorithm. The proposed algorithm achieved highest accuracy using single-channel EEG, till date. It achieved an accuracy of 82.35% with mean kappa of 0.64. For validation, 5-fold cross validation has been used. Furthermore, the presented paper has low computational cost, which makes GB-based model suitable for real time neonatal sleep wake classification.